# Reconstruction of Contour Lines During the Digitization of Contour Maps to Build a Digital Elevation Model

*Aroj Subedi[1], Pradip Ganesh[2], Sandip Mishra[3]*
1,2,3- Computer Science and Engineering Department, Sikkim Manipal Institute of Technology, Majitar, Rangpo, Sikkim, India.(arojsubedi@gmail.com)

**Abstract:** Contour map has contour lines that are significant in building a Digital Elevation Model (DEM). During the digitization and pre-processing of contour maps, the contour line intersects with each other or break apart resulting in broken contour segments. These broken segments impose a greater risk while building DEM leading to a faulty model. In this project, a simple yet efficient mechanism is used to match and reconnect the endpoints of the broken segments accurately and efficiently. The matching of the endpoints is done using the concept of minimum Euclidean distance and gradient direction while the Cubic Hermite spline interpolation technique is used to reconnect the endpoints by estimating the values using a mathematical function that minimizes overall surface curvature resulting in a smooth curve. The purpose of this work is to reconnect the broken contour lines generated during the digitization of the contour map, to help build the most appropriate digital elevation model for the corresponding contour map.

## 1. INTRODUCTION

This chapter presents a general discussion related to the contour maps, digitization of contour maps, pre-processing, reconnection of broken contour lines, building a Digital Elevation Model (DEM), and applications of the generated DEM.

The contour line is one of the major features available on the topographic map. A map consisting of only contour lines can be generated which is known as a contour map. A contour map is a topographic map having contour lines often in a monochromatic brown color that join points of equal elevation on the surface of the land above or below a reference surface. Contour lines are also called isolines and the relative spacing of the lines in the contour map indicates the relative slope of the surface. Thus, contours make it possible to represent the height of mountains, depth of an ocean, or a lake, and steepness of slopes on a two-dimensional surface. For steep places, the lines are close together and the elevation is changing rapidly while for the flat places, the elevation doesn't change much.

The analog maps are reproduced in the digital form through a process of digitization. The digitization of the contour map is done to trace the map to a digital form for two major purposes, for the features of the map to have a proper locational identity and the

This research did not receive any specific grant from funding agencies in the public, commercial, or not-for-profit sectors.

other, to create a digital elevation model as the two-dimensional surface imposes visualization problem for the users. Digitization is an expensive and time-consuming process and can be done using two different methods: manual digitization and heads up digitization (i.e raster scanning using optical scanners). The important aspect of digitization is always the accuracy of the digitized product. The lines and the points on the map represent a considerable area on the surface of the earth, so, the lines and points have to be presented accurately, or else it might result in different kinds of tension.

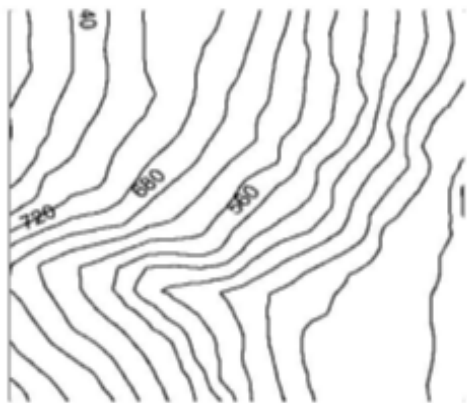

**Figure 1. Sample contour map**
*Ref: Sikkim State Disaster Management Authority, Government of Sikkim*

Due to the information layers such as the coordinate grid and the toponymy present in the maps, the digitization of the contour maps may result in the broken contour lines/segments and the pre-processing phase as well may result in the unwanted pixels as noise. The broken contours may result in the inaccurate placement of the features thus, not matching the image and the digitized image and further resulting in the inaccurate depiction of the DEM compared to that of the original surface of the land, therefore leading to a faulty model. Thus, a proper advanced strategy has to be used to reconnect the broken segments of the contour line with the utmost accuracy and efficiency and to remove the unwanted noisy pixels.

The elevation values present in the map can be generated during the pre-processing phase or after the reconnection of the contour lines phase is complete. The elevation value along with the generated DEM data can be processed further using different techniques and software to generate a proper DEM model.

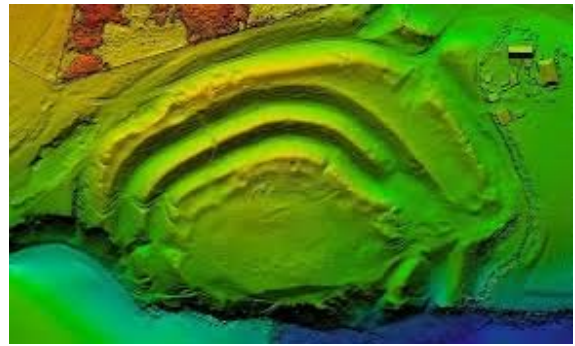

**Figure 2. Sample DEM**
Ref: https://theconstructor.org/

The generated DEM model can then be used to determine the landslide possibility within a geographical region, to analyze changes within a topographic region within a certain period of time as well as has several other engineering and geographical application.

This paper deals with the digitization of contour maps, extraction of features such as elevation value from the contour maps so that it can act as a basis for constructing a Digital Elevation Model (DEM). The primary focus of this work is to reconnect the broken contour lines that are generated during the digitization process of the contour maps so, that the digitized contour map overlaps with the original map and also the features of the map have the correct locational identity corresponding to the original map, therefore resulting in accurate 3-D depiction in the form of DEM of the corresponding map.

## 2. RELATED WORK

The continuity-based contour reconstruction method [10, 11] uses different characteristics of a contour line such as topological and geometrical properties to identify the broken contour endpoints. Arrighi et al. [12] proposed the reconstruction of the broken contour line using the minimum point pair method to create a complete contour line. Most of the work carried out for the contour reconstruction is performed by matching the endpoints using the distance and the auxiliary direction. Ghircoias et al. [3] proposed the use of local information, i.e. Euclidean distance between the extreme points of the curve and their directions combined for the joint decision. The proposed solution was able to achieve a contour reconstruction accuracy of 83.35%. Hancer et al. [1] considered both opposite as well as parabolic directions during the endpoint matching and minimized the drawbacks of the conventional methods of reconnection of broken segments. The efficiency and the accuracy of reconnection were also improved significantly, but it didn't include the reconstruction method based on the elevation values of the contour lines. Samet et al. [4] proposed a technique to resolve and remove the crossed points effectively by using masks and a combination of distance and direction criteria for reconnection procedure. The problem with this kind of approach arises in the case of a larger broken contour line area and may result in an incorrect matching of broken segments thus, giving rise to intertwined contour lines. Pradhan et al. [2] devised a different knowledge-based efficient reconnection technique based on the concept of a leech, water flow, and wiper. The implemented techniques efficiently reconnected the broken contour lines and required very less knowledge regarding the state of other contour lines and prevented the crossover between the different broken lines, though the search space could've been reduced significantly by including the knowledge of the spline curve. Amenta et al. [14] proposed an approach based on Delaunay Triangulation and Voronoi Diagrams which use the concept of the medial axis of a curve. The problem with this approach is insufficient information and unsolved gaps. The introduction of a more automatic method of reconstruction of broken contour segments was always missing in most of the proposed methods. Pouderoux and Spinello [16] proposed an automatic approach to reconstruct the broken contour lines based on the gradient orientation field generation and perfect matching algorithm to fill the gaps. The downside of this approach is the processing time. Chengming et al. [13] proposed a reconstruction method for the broken contour lines based on the reference line determined using similar and completely closed contours. The node densification was conducted on the broken contour lines to improve the identification accuracy of the reference line, and the discrete Fréchet distance was used to select a reference line and perform the reconstruction process. The result of this approach showed a good reconstruction for both general and complex broken contours, with good universality. To build the digital elevation model from the contour maps, the elevation value feature present in the contour map has to be extracted. Luyang et. al. [5] proposed a technique to locate the text string box and to identify the external line vector incident on the box but didn't address the identification of strings crossed over by the contour lines. Xie et al. [6] presented an efficient algorithm to label the contour lines and to convert contour maps into digital elevation maps.

## 3. PROPOSED WORK

A few of the challenges faced while converting a contour map to a 3-D model are:

a. Identifying terminal and crossed points. Removing crossed points and devising a suitable technique to reconnect the broken contour line segments.

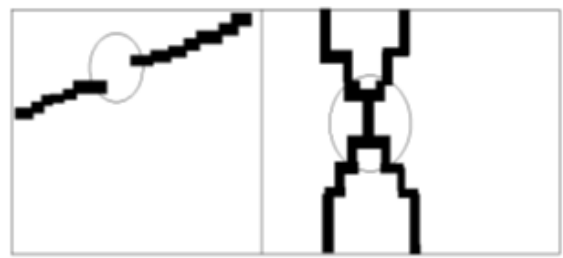

**Figure 3(a)**      **Figure 3(b)**
*Figure 3(a). Terminal points in a broken contour segment Figure 3(b). Crossed points in a contour segment*

b. Character extraction and recognition to generate the elevation value. Some digits like 0 and 8 don't have horizontal and vertical lines and endpoints as well.

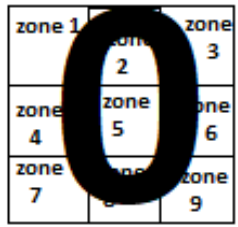

**Figure 4. Zoning of digit 0.**

c. Recognizing characters that are intersected/crossed by contour lines.

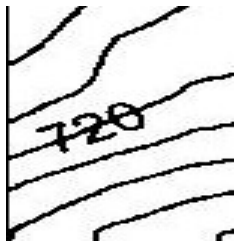

**Figure 5. Elevation value crossed by a contour line**

The proposed work aims at tackling the challenges faced while converting contour maps to a 3D model by using the following different techniques at different steps.

To detect the crossed points, a 3 by 3 mask is applied to the whole image and since a contour can move only in one direction, the pixels moving in two or more directions are deleted.

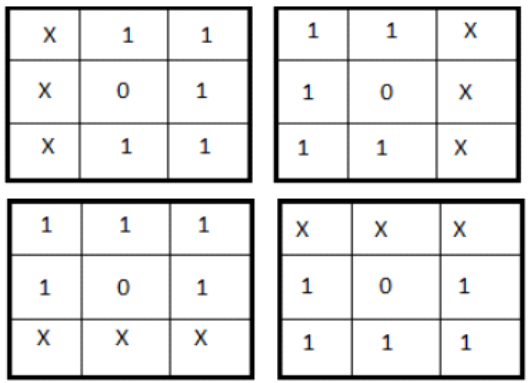

**Figure 6. 3x3 mask applied to detect crossed points**

To detect the endpoints, filters are used. A search window of 11 * 11 pixels is established to detect the endpoints in the region. To match the endpoint of a broken contour with the other endpoint, a combination of minimum Euclidean distance and direction of the pixel computed using gradient operator is used and the matching endpoints are reconnected through the Cubic Hermite spline interpolation technique.

The Cubic spline C is defined as

$$C_i = 3a_{i-1}h^2 + 2b_{i-1}h + c_{i-1}, \quad i = 1,2,\ldots, n\text{-}1 \quad (1)$$

Where,

$$h = x_i - x_{i-1}; a_i = \frac{M_{i+1} - M_i}{6h}; b_i = \frac{M_i}{2}$$

$$c_i = \frac{y_{i+1} - y_i}{h} - \left(\frac{M_{i+1} + 2M_i}{6}\right)h$$

And,

$$d_i = y_i$$

Also,

$$C_{i+1} = 3a_i h^2 + 2b_i h + c_i$$
$$= M_i + 4M_{i+1} + M_{i+2}$$
$$= 6\left(\frac{y_i - 2y_{i+1} + y_{i+2}}{h^2}\right) \quad \text{where } i = 1,2,3,\ldots,n\text{-}1$$

The Spline is a curve that is used to reconnect two or more points by drawing smooth curves. The points are numerical data and the weights consisted of splines are the coefficients on the cubic polynomials used to interpolate the data. These coefficients bend the line so that it passes through each of the data points without any break in continuity. The s=spline(x,y,xq) inbuilt function returns a vector of interpolated values s corresponding to the query points in xq. The values of s are determined by cubic spline interpolation of x and y [8]. Spline chooses the slopes differently, namely to make even $S''(x)$ continuous so that the produced result is smoother.

A text-string box is located and an external line vectors incident on the box is identified. The binary image is divided into 9 zones and the number of foreground (black) pixels in each zone is counted to find out the zone with specified foreground pixel value to detect a specific digit. For example, digit 0 has zero foreground pixel value in zone 5.

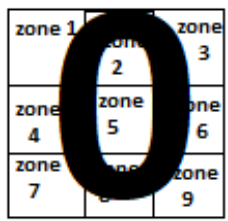

**Figure 7. Feature Extraction of digit 0.**

To label the contour lines and to build DEM, the concept of point extrema, triangulation, and generation of height values using the bilinear interpolation technique is used. The DEM can also be generated automatically using software such as ArcGIS.

## 4. DESIGN

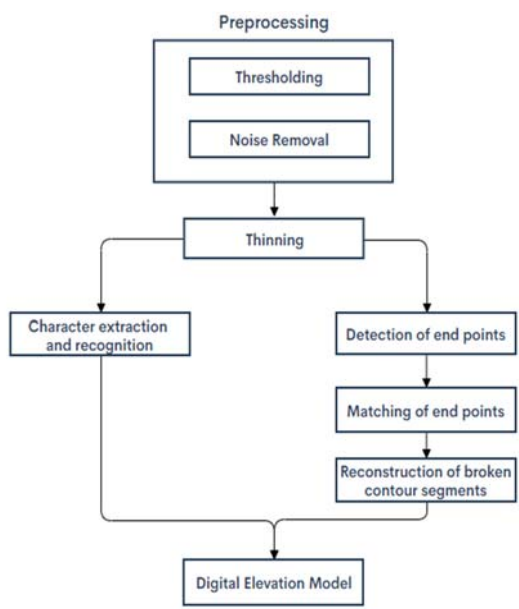

**Figure 8. Design flowchart of the proposed system**

To generate the Digital Elevation Model (DEM) from the contour maps, the following steps are followed:

### 4.1 Pre-processing:

To enhance raw images. The image pre-processing technique mainly deals with the thresholding and noise removal operations to perform morphological operations and to implement a general algorithm on the image acquired.

#### 4.1.1 Thresholding operation:

Used to extract useful information while reducing the background noise. Thresholding operation is performed using the Histogram shape-based method to extract only the essential information.

#### 4.1.2 Noise Removal operation:

Used to remove unwanted pixels as distortions. A median filter is used to remove noise such as Gaussian noise and Salt and Pepper noise.

### 4.2 Thinning operation:

Used to remove selected foreground pixels from binary images and is used for skeletonization. Zhang-Suen's algorithm [7] is used to convert the thick pixel image into a single-pixel thick image.

### 4.3 Detection of endpoints:

Terminal points of the broken contour segments are detected using filters over a window of certain pixels.

### 4.4 Matching of endpoints:

The detected endpoints are matched with each other using the concept of Euclidean distance and gradient direction to proceed with the process of reconstruction of the contour line.

The Euclidean distance between two points $(x_1, y_1)$ and $(x_2, y_2)$ is defined as

$$ed = \sqrt{(x_1 - x_2)^2 + (y_1 - y_2)^2} \qquad (2)$$

*4.5 Reconstruction of broken contour segments:*

The matched endpoints are reconnected to reconstruct broken segments using the Cubic Hermite spline interpolation technique. The curve fitting process with the cubic spline interpolation technique results in the consistency and efficiency of the spline. The Cubic Hermite interpolation method estimates values using a mathematical function that minimizes overall surface curvature, resulting in a smooth surface that passes exactly through the input endpoints.

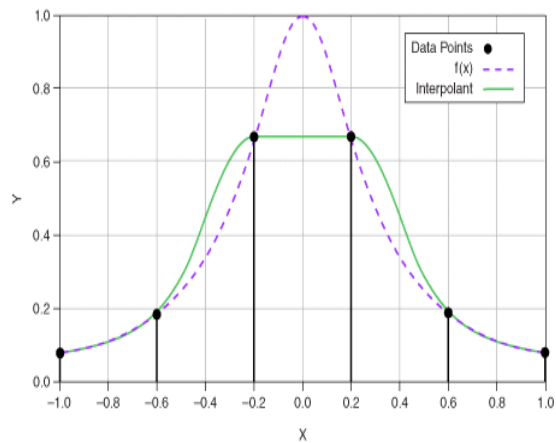

**Figure 9. Cubic Hermite spline interpolation for 11 data points for f(x)**
*Ref: UZH - Physik-Institut*

*4.6 Character extraction and recognition:*

To identify the location of the text string in the map and recognize the characters to generate the elevation value.

*4.7 Digital Elevation Model (DEM):*

To generate a 3D model from the DEM data in 3D space.

# 5. IMPLEMENTATION DETAILS

*5.1 Pre-processing of the image read*
*5.1.1 Thresholding*

With the help of histogram information, a thresholding operation was performed.

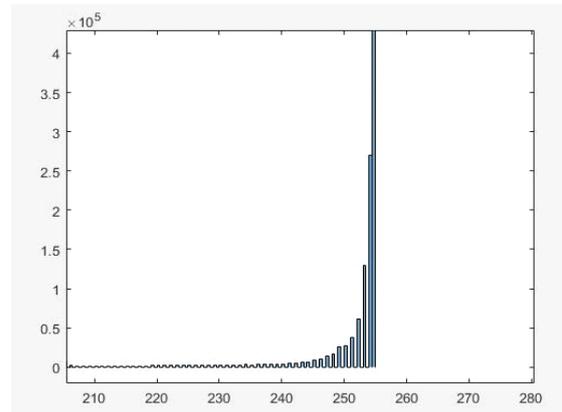

**Figure 10. Histogram for the image read**

---

Algorithm 1: Thresholding algorithm

*Given*: An image 'img'
*Objective*: To convert the image to a binary image for easier analysis
*Procedure*:
1. Generate histogram values to find the spread of the pixel values.
2. Compute the mid-point of the spread as 'm'.
3. Compute the number of rows of a pixel as 'r' and columns of pixel as 'c'
4. for i ← 1 to r
   4.1. for j ← 1 to c
      4.1.1. if img(i,j) > m
         4.1.1.1. img(i,j) = 0
      4.1.2. else
         4.1.2.1. img(i,j)=255
      end if
   end for
end for

---

### 5.1.2 Noise Removal operation

Salt and pepper noise generated after binarization was removed with the help of the median filter by using the inbuilt function *medfilt2()*. Different colored pixels in the binary image as Gaussian noise was also removed using the inbuilt function *medfilt2()*.

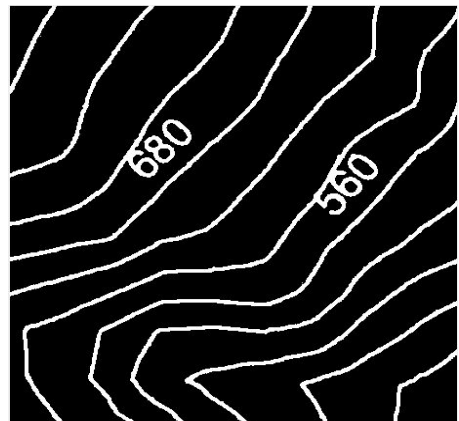

**Figure 12(a). Before thinning thick pixel image**

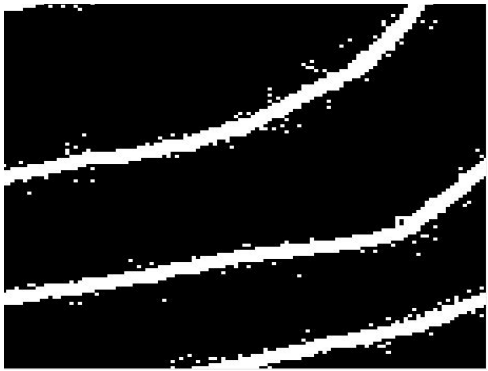

**Figure 11(a). Salt and pepper noise**

The Zhang-Suen Thinning algorithm [7] is a 2-pass algorithm. Two sets of checks are performed in each iteration to remove pixels from the image. The first set removes from the southeast (bottom right) corner of the image, and the second set removes from the northwest (top left) corner of the image.

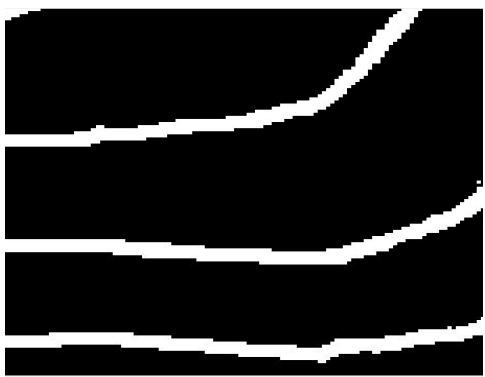

**Figure 11(b). After removal of salt and pepper noise**

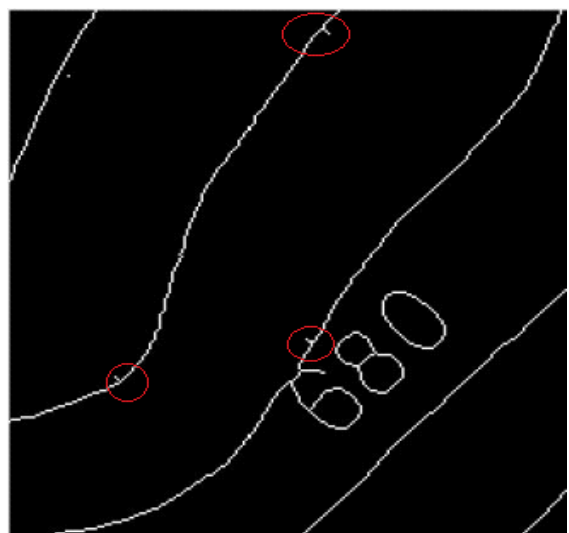

**Figure 12(b). Single-pixel thick image produced by bwmorph function**

### 5.2 Thinning Operation

The inbuilt morphological operator function *bwmorph()* produced unnecessary protrusions and to remove these protrusions, the thinned image required additional pre-processing. Instead of bwmorph morphological operator function, the thinning algorithm as proposed by Zhang-Suens [7] was implemented and produced satisfactory results.

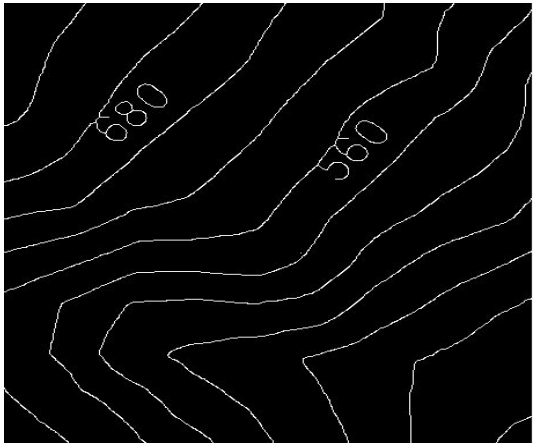

**Figure 12(c). Single-pixel thick image produced after thinning using the Zhang-Suen algorithm [7]**

*5.3 Detection of Endpoints*

Terminal points of the broken contour segments are detected.

---
Algorithm 2: Terminal points detection algorithm
*Given*: A binary image 'img', number of rows of a pixel 'r' and columns of pixel 'c'
*Objective*: To detect the terminal points i.e. endpoints in a contour line
*Procedure*:
1. for i ← 2 to r-1
    1.1. for j ← 2 to c-1
        1.1.1. S = sum of the pixels in the
               8-neighborhood of img(i.j)
        1.1.2. if S == 1
            1.1.2.1. Terminal point detected
            end if
        end for
    end for

---

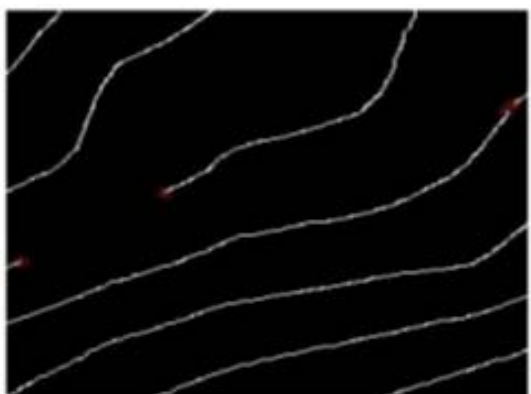

**Figure 13. Endpoints detected**

*5.4 Matching of endpoints*

The detected endpoints are matched with each other to facilitate the reconstruction process. The gradient features of the terminal points along the x and y-axis as $G_x$ and $G_y$ are evaluated using the 'Sobel' operator and also the gradient direction is evaluated to help in the reconstruction purpose.

The gradient direction for the terminal points (x,y) having gradient features $G_x$ and $G_y$ is computed as

$$\alpha(x, y) = \tan^{-1}\left(\frac{G_y}{G_x}\right) \qquad (3)$$

| X – Direction Kernel | | | Y – Direction Kernel | | |
|---|---|---|---|---|---|
| -1 | 0 | 1 | -1 | -2 | -1 |
| -2 | 0 | 2 | 0 | 0 | 0 |
| -1 | 0 | 1 | 1 | 2 | 1 |

**Figure 14. Sobel operator to determine $G_x$ and $G_y$**

*Steps for matching endpoints*

Search space of say 11 * 11 pixels is used in the image to detect the endpoints in the region.

---
Algorithm 3: Gradient direction checking algorithm
*Given*: Gradient direction 'gd' of the end point detected
*Objective*: To assign a parameter value 'dir' to aid in end point matching
*Procedure*:
1. if gd > 0
    1.1. set dir = 1
2. else
    2.1. set dir = 0
end if

---

A detected endpoint is matched with the other endpoint using the concept of minimum Euclidean distance and gradient direction.

| Algorithm 4: Gradient direction checking algorithm |
| --- |
| *Given*: Parameter value 'dir' of the endpoint
*Objective*: To match the detected endpoint with another appropriate endpoint
*Procedure*:
1. if dir == 1
   1.1. select detected endoint with gradient direction < 0
2. else
   2.1. select detected endpoint with gradient direction > 0
end if |

Any endpoint that doesn't match is saved and checked against other similar endpoints at the end.

### 5.5 Reconstruction of broken contour segments

The matched endpoints are reconnected using smooth polynomial functions called splines. For reconnection, the Cubic Hermite spline interpolation technique is used.

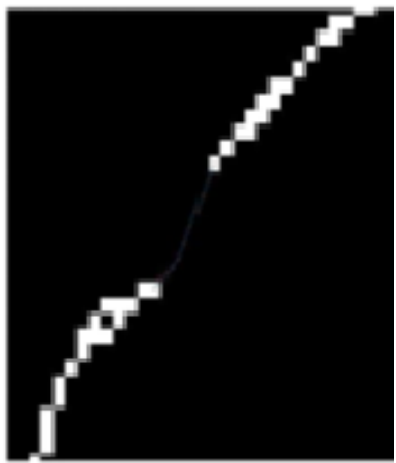

**Figure15(a). Interpolated using the Cubic Hermite spline**

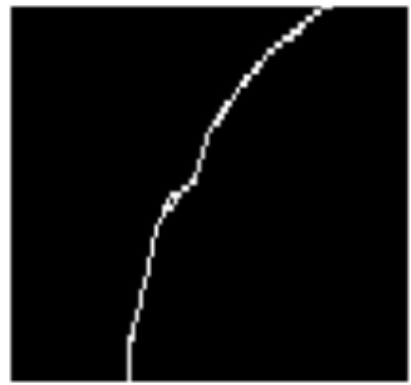

**Figure15(b). Endpoints reconnected by adding foreground pixels along the splines.**

The project is implemented in MATLAB version R2015b environment by importing several inbuilt functions and writing the raw code to achieve the desired objective.

### 6. EXPERIMENTAL RESULTS

This chapter presents a general discussion related to the results obtained after the implementation of the project.

The histogram values were generated to find the spread of the pixel values of the contour map and the threshold as the mid-point of the spread was found to be 243.

The implemented code was able to detect a total of 6 endpoints having co-ordinate values (424,248), (428,245), (438,461), (478,95), (480,438) and (520,32) having respective gradient direction 45, -90, 90, 45, -135 and -135 degrees. The pixels at co-ordinate values (424,248), (438,461) and (478,95) were matched with one of the pixel present at co-ordinate values (428,245), (480,438) and (520,32).

*Case 1:*
When the endpoints are detected and matched within the search window.

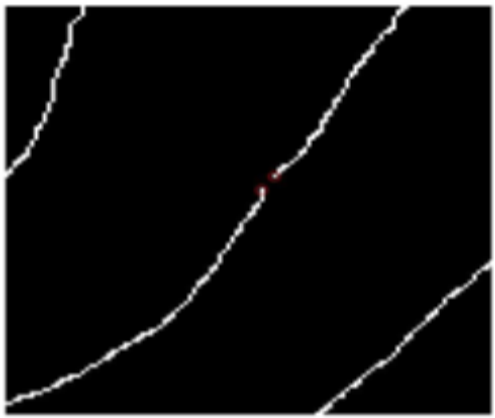

**Figure 16. Endpoints detected and matched**

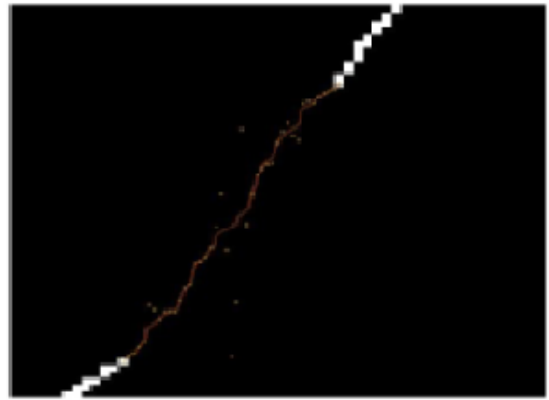

**Figure 18. Spline generated path for reconnection of pixels at co-ordinate values (438,461) and (480,438)**

The above figure shows endpoints detected at the co-ordinate value (424,248) and (428,245). The endpoints are matched using the concept of minimum Euclidean distance which is 5 units and the gradient direction which is opposite.

*Case 2:*

When the endpoints are not detected within the search window, an entry for the endpoint is maintained and checked against other endpoints at the end. An endpoint is selected and other matching endpoints are picked using the concept of opposite gradient direction. The matching endpoint is then selected using the concept of minimum Euclidean distance.

The endpoint (438,461) is matched with (480,438) and the endpoint (478,95) is matched with (520,32).

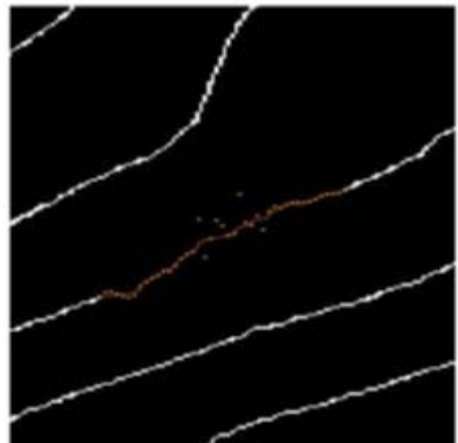

**Figure 19. Spline generated path for reconnection of pixels at co-ordinate values (478,95) and (520,32)**

The foreground pixels are added along the spline generated path to complete the reconnection process to reconstruct the broken segments.

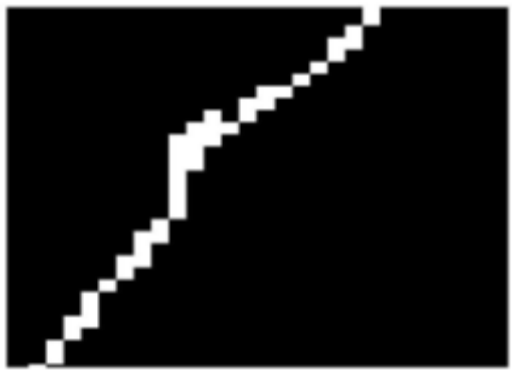

**Figure 17. Reconnected endpoint at co-ordinate values (424,428) and (428,245)**

## 7. CONCLUSION AND FUTURE WORK

The use of histogram information helped to process the raw image sharply and the implementation of the remaining steps happened to be easier. The median filters were able to remove the noise and the thick pixel image was converted into a single-pixel image using the Zhang-Suen algorithm [7]. The concept of Euclidean distance and gradient direction was able to match the detected endpoints. The use of only two features for matching could cause problems if many endpoints are detected. The character recognition step is tricky if the characters are overlapped with the contour lines. The concept of Euclidean distance and gradient direction is though applicable, some new features have to be taken into account while trying to match the endpoints for the reconnection process.

More pre-processing steps can be performed to sharpen the contour map read and to nullify the unwanted distortions as noise. The matching of endpoints can be made more accurate by considering other directional features. The bidirectional parabola control algorithm proposed in [15] can be used to reconstruct the gaps having acute angle parabola. A reconstruction method proposed in [13], based on reference contour lines selected using discrete Fréchet distance can be used to reconstruct larger broken contour lines. The broken segments can be reconstructed by generating smooth curves using an efficient interpolation technique. The extraction and identification of elevation values from the contour map can also be done using efficient techniques. The development of an automatic technique for contour line reconstruction can also be considered as one of the future directions. A significant amount of work can be done to increase efficiency, and to reduce the complexity of the method.